\pdfoutput=1

\documentclass[11pt]{article}

\usepackage{EMNLP2022}

\usepackage{times}
\usepackage{latexsym}
\usepackage{multicol}
\usepackage{multirow}
\usepackage{colortbl}

\usepackage{wrapfig}
\usepackage{graphicx}
\usepackage{cleveref}
\usepackage{bm}
\usepackage{svg}
\usepackage{balance}

\usepackage[T1]{fontenc}

\usepackage[utf8]{inputenc}
\usepackage{soul}
\usepackage{arydshln}
\usepackage{booktabs}
\usepackage{microtype}

\title{Iteratively Prompt Pre-trained Language Models for Chain of Thought\\}

\author{Boshi Wang, \ \ Xiang Deng \and Huan Sun \\
        The Ohio State University, Columbus, OH \\
        \texttt{\{wang.13930,deng.595,sun.397\}@osu.edu}}

\begin{document}
\maketitle
\begin{abstract}
While Pre-trained Language Models (PLMs) internalize a great amount of world knowledge, they have been shown incapable of recalling these knowledge to solve tasks requiring complex \& multi-step reasoning. Similar to how humans develop a ``chain of thought'' for these tasks, how can we equip PLMs with such abilities? In this work, we explore an iterative prompting framework, a new prompting paradigm which progressively elicits relevant knowledge from PLMs for multi-step inference. We identify key limitations of existing prompting methods, namely they are either restricted to queries with a single identifiable relation/predicate, or being agnostic to input contexts, which makes it difficult to capture variabilities across different inference steps. We propose an iterative context-aware prompter, which addresses these limitations by learning to dynamically synthesize prompts conditioned on the current step's contexts. Experiments on three datasets involving multi-step reasoning show the effectiveness of the iterative scheme and the context-aware prompter design.\footnote{Our source code is available at \url{https://github.com/sunlab-osu/IterPrompt}.}

\end{abstract}

\section{Introduction}
\label{sec:intro}

\begin{figure}[t]
  \centering
    \includegraphics[width=0.48\textwidth]{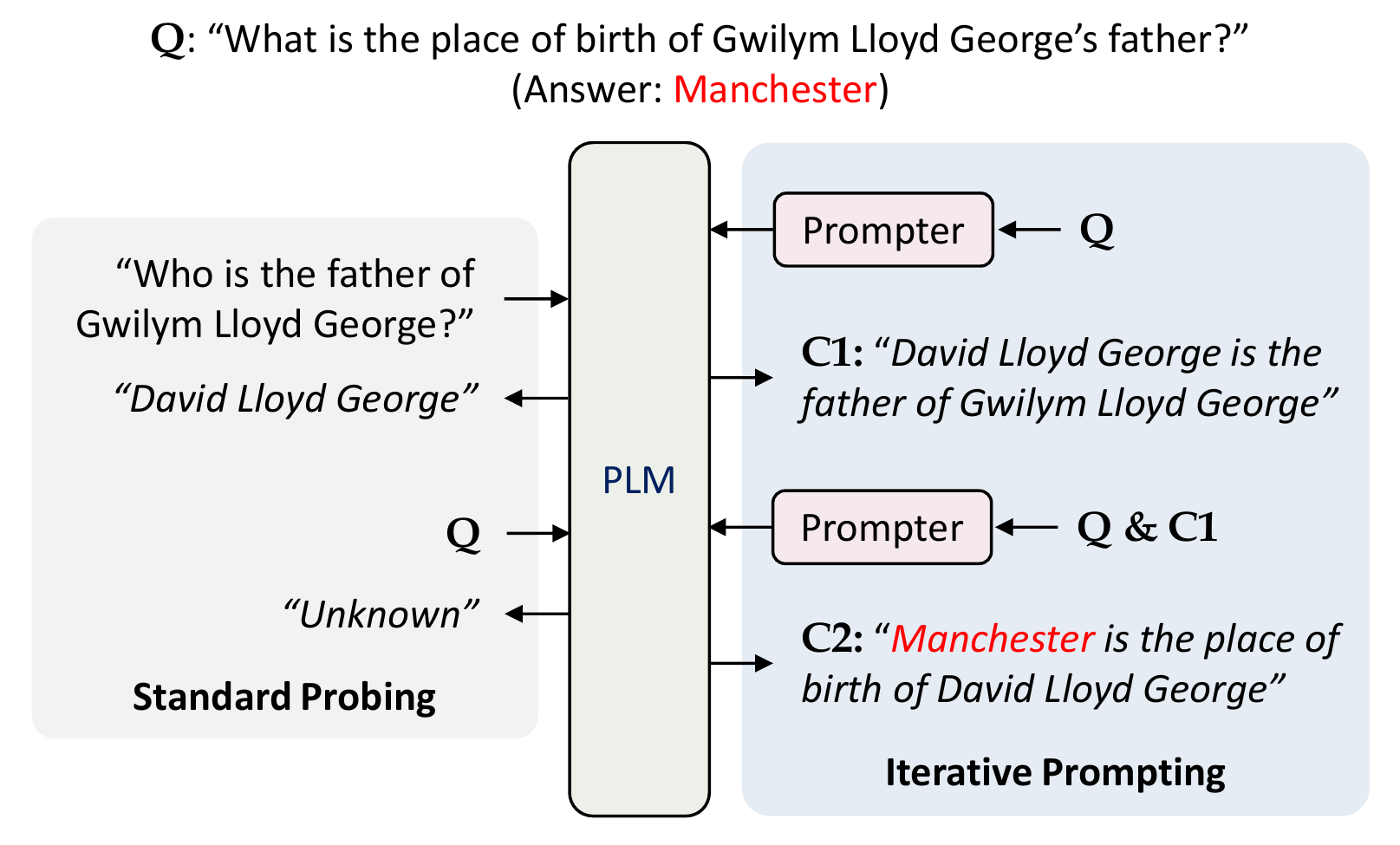}
    \caption{Our Iterative Prompting approach (on the right), compared with Standard Probing (on the left). In Standard Probing, a question is directly fed to the PLM to output the final answer, which could work for simple factual questions but fails for complex questions that require multi-step reasoning. In contrast, we augment the PLM with a Prompter, which learns to iteratively prompt the PLM to recall a series of knowledge and derive a ``chain of thought''.}
    \vspace{-1em}
\label{fig:pipeline}
\end{figure}

Humans can develop a ``chain of thought'' for complex decision making. For example, when asked the question (\textbf{Q}) shown in Figure \ref{fig:pipeline}, which involves {composition}, an important type of multi-step reasoning, humans apply two consecutive steps to derive the final answer: 1) find ``father'' of the topic entity ``Gwilym Lloyd George'' (\textbf{C1}); 2) find ``birthplace'' of the entity returned in the first step (\textbf{C2}). 

Recently, large-scale pre-trained language models (PLMs) have been shown capable of internalizing a great amount of simple factual knowledge such as \textbf{C1} and \textbf{C2}, yielding competitive performance on a range of knowledge-intensive tasks without resorting to any external knowledge source \cite{petroni2019language, shin-etal-2020-autoprompt, zhong2021factual, roberts2020much, lee2020language}. However, work such as \cite{talmor2020olmpics, kassner2020pretrained, rae2021scaling} reveals that PLMs face difficulties in \textit{complex, multi-step} reasoning. For example, they struggle with answering complex questions like \textbf{Q} without using external sources, no matter whether they are fine-tuned based on QA pairs or simply prompted to produce the answer (where even if they have memorized \textbf{C1} and \textbf{C2}).
%
%

In this paper, we study the following question: How to shepherd a PLM to \textit{recall a series of stored knowledge} (e.g., \textbf{C1} and \textbf{C2}) that is necessary for multi-step inference (e.g., answering \textbf{Q}), analogous to how humans develop a ``chain of thought'' for complex decision making?  

A direct way would be to fine-tune the PLM to generate the series of knowledge all at once (assuming such supervision is available), but soon one realizes the practical issue in this approach: PLMs which internalize a great amount of knowledge are inevitably large in scale, and fine-tuning all their parameters would become more and more costly as they keep scaling up. There is also the concern that fine-tuning PLMs may interfere with their implicit knowledge storage, a phenomenon observed in \cite{wang-etal-2021-generative} which is more generally related to the catastrophic forgetting problem of deep learning models \cite{mccloskey1989catastrophic, kirkpatrick2017overcoming, howard-ruder-2018-universal}. Therefore, lightweight methods such as prompting \cite{liu2021pre} which keep a PLM's parameters intact would be preferable for our purpose of eliciting knowledge. However, we find that no matter whether it is fine-tuned or prompted to generate the series of knowledge all at once, the PLM tends to lose its ``chain of thought'' during the process, generating irrelevant facts or suffering from hallucination. 


Motivated by the iterative nature of multi-step reasoning problems, we explore an \textit{iterative prompting} framework in this paper, 
which elicits knowledge from PLMs step by step for a given inference task. We have two desiderata in iterative prompting: (1) At different inference steps, the prompts need to focus on different components of the complex query. (2) The prompts should appropriately integrate knowledge gathered in previous steps into the current step; for instance, during the 2nd step in the example in Figure \ref{fig:pipeline}, the prompts need to combine the entity ``David Lloyd George'' (from knowledge recalled in the 1st step) with the unresolved part ``What is the place of birth of'' in the query. 

A natural thought is to directly apply existing prompting methods in an iterative fashion. Unfortunately, their prompts are either restricted to queries with a single, identifiable relation/predicate \cite{jiang2020can, petroni2019language, zhong2021factual, shin-etal-2020-autoprompt, qin-eisner-2021-learning}, or being agnostic and insensitive to step-wise inputs \cite{lester-etal-2021-power, li-liang-2021-prefix, brown2020language}, and hence not ideal for our desiderata.

We design a novel iterative prompting method towards that end. We augment the PLM with an \emph{iterative Context-Aware Prompter}, a model which learns to \textit{dynamically synthesize prompts} based on the current step context. At each step, the Prompter learns to process the query and previously gathered evidence, and composes a prompt which steers the PLM to recall the next piece of knowledge. Like other prompting methods, the PLM is kept fixed throughout the learning process. In addition, as the PLM size increases, the number of trainable parameters in our method scales comparably with or slower than previous prompting methods.

We conduct experiments on three datasets involving multi-step reasoning, including two recent multi-hop Question Answering datasets: 2WikiMultiHopQA \cite{ho2020constructing} and R4C \cite{inoue-etal-2020-r4c}, and a scientific dataset \cite{talmor2020leap} for reasoning over taxonomic relations. Our experimental results show (1) effectiveness of the iterative scheme; 
(2) our proposed Context-Aware Prompter design outperforms existing prompting methods by notable margins; (3) quantitative and qualitative analysis which reveal the faithfulness of our learned prompter.

\section{Methodology}
In this section, we first formalize our problem setup (\S\ref{sec:probsetup}), and then introduce our iterative prompting framework (\S\ref{sec:iterprompt}), followed by our context-aware prompter design (\S\ref{sec:prompter}) which addresses key limitations of previous prompting methods when applied in this iterative scheme.

\subsection{Problem Setup}
\label{sec:probsetup}
Given a complex query $q$, our goal is to drive a PLM $\mathcal{M}$ to recall a sequence of simple knowledge statements $C_q=[c_1,...,c_{n_q}]$ which is sufficient for deciding the response to $q$. In particular, we focus on developing prompting methods, where the parameters of $\mathcal{M}$ are fixed and we aim to construct prompt $T$ which steer $\mathcal{M}$ to recall $C_q$. Note that here we treat $T$ as a \textit{variable}, which may or may not depend on other variables based on different modelings. Writing $\mathcal{M}(T)$ as $\mathcal{M}$ augmented with prompt $T$, our training objective is to learn how to find $T$ which could maximize the log-likelihood

$$
    \mathcal{L}(T)=\sum_{i=1}^N \log P(C_{q_i}|q_i; \mathcal{M}(T))
$$ 

\noindent with a set of training data $\{q_i, C_{q_i}\}_{i=1}^N$. 

Our formulation here is general and applicable to all prompting-based methods, where the settings in previous work such as \cite{zhong2021factual, shin-etal-2020-autoprompt, lester-etal-2021-power, li-liang-2021-prefix, qin-eisner-2021-learning} correspond to the reduced case where $|C_q|=1$ for any query $q$. In our experiments, we also consider PLM fine-tuning, in which case there's no prompt $T$ in the pipeline, and instead the parameters of $\mathcal{M}$ are optimized.

\subsection{Iterative Prompting Framework}
\label{sec:iterprompt}
Inspired by the sequential nature of multi-step inference tasks, we approach the problem in an iterative way:
$$
    P(C_q|q; \mathcal{M}(T)) = \prod_{j=1}^{n_q} P(c_j|q, c_1,..., c_{j-1}; \mathcal{M}(T))
$$
where at each step $j$, $\mathcal{M}(T)$ recalls the next piece of knowledge $c_j$ conditioned on the query $q$ and all previously gathered knowledge $c_1, ..., c_{j-1}$ (concatenated with $q$).

\subsection{Context-Aware Prompter}
\label{sec:prompter}
Previous prompting methods which take single-relation inputs clearly fail to apply in this iterative setting due to the complexity of the input context $q, c_1,..., c_{j-1}$. Task-level prompting methods such as Prompt-Tuning \cite{lester-etal-2021-power} and Prefix-Tuning \cite{li-liang-2021-prefix} are applicable here, where $T$ is treated as a \textit{static} parameter. However, as described earlier, this modeling is not ideal for $T$ to fully capture variabilities across different inference steps. In this work, we model $T$ as the output of our \textit{Prompter}, a learnable function mapping $f_W$ which dynamically synthesizes $T$ \textit{w.r.t.} the current step input context:
$$
    T = f_W(q, c_1,..., c_{j-1}), \forall j
$$

\begin{figure}[t]
  \centering
    \includegraphics[width=0.95\linewidth]{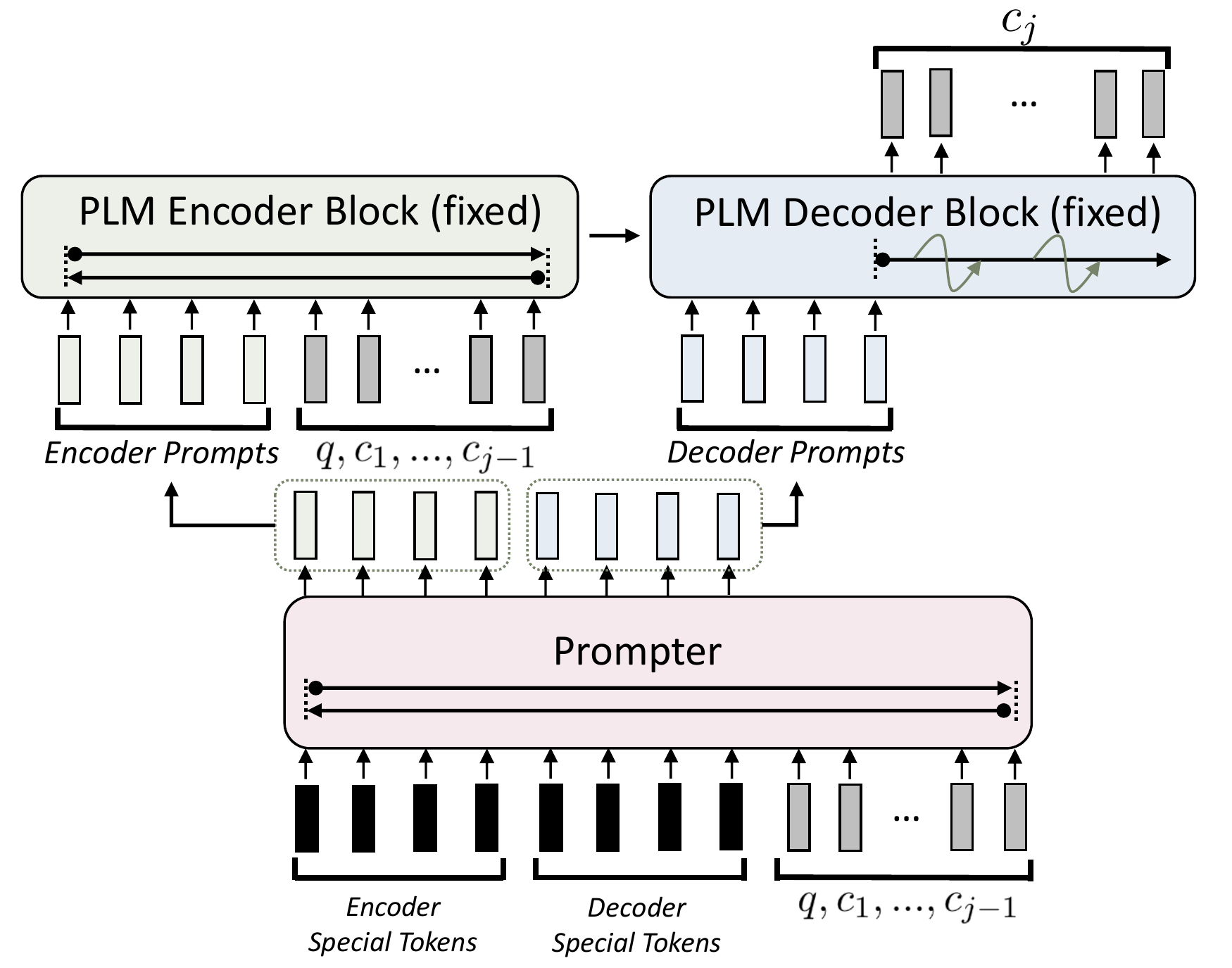}
  \caption{Our context-aware prompter design. The prompter contextualizes a set of special tokens \textit{w.r.t.} the current step context $q, c_1,..., c_{j-1}$ to get the resulting prompts, which steers the PLM to recall the next piece of knowledge $c_j$.}
\label{fig:prompter}
\end{figure}

\noindent \textbf{Prompter Instantiation.} While there are many plausible design choices for the Prompter $f_W$, here we instantiate it with a transformer-based language model (shown in Figure \ref{fig:prompter}). The prompts are designed to be contextualizations (by the Prompter) of a set of special tokens \textit{w.r.t.} the current step input context, linearly projected into the PLM's embedding space by a trainable matrix (omitted in the figure due to space limit). In this work, we adopt an Encoder-Decoder PLM and use prefix-prompts in the implementation; hence we have prompts that are prepended to both the PLM's encoder inputs and decoder inputs. Note that our design could be easily adapted to other types of PLMs (e.g., encoder-only/decoder-only models) and different prompt positionings (e.g., infix, postfix).

\noindent \textbf{Comparison with Prompt/Prefix-Tuning.} Both Prompt-Tuning \cite{lester-etal-2021-power} and Prefix-Tuning \cite{li-liang-2021-prefix} model the prompt $T$ as a context-agnostic parameter. In Prompt-Tuning, $T$ has the same identity as in our approach which is a set of virtual input tokens (\textit{Encoder Prompts} \& \textit{Decoder Prompts} in Figure \ref{fig:prompter}). In Prefix-Tuning, $T$ is modeled to be the set of activations (keys \& values in the transformer attention blocks) of the virtual prompt tokens across all PLM layers. Let $D$ be the embedding dimension of the PLM, $h$ be the number of layers in the PLM, $d$ be the embedding dimension of the Prompter ($d \le D$), and $l$ be the length of the prompt tokens (both encoder \& decoder prompts). Then the number of trainable parameters is $\Theta(d\cdot (D+l))$ for our proposed method, $\Theta(l\cdot D)$ for Prompt-Tuning and $\Theta(l\cdot h\cdot D)$ for Prefix-Tuning. It can thus be seen that our proposed method scales comparatively with Prompt-Tuning, slower than Prefix-Tuning, and overall maintains a manageable amount of trained parameters as the PLM scales up (which increases $D$ and $h$).


\noindent \textbf{Continuous v.s. Discrete Prompts}. While modeling $T$ as discrete tokens in the PLM's vocabulary could increase the readability of the prompts, a discrete space is much less expressive than its continuous counterpart, and optimization over a discrete space could be highly inefficient. Also, despite being inside the vocabulary, the searched discrete prompts could still have low interpretability as seen by the given examples in \cite{shin-etal-2020-autoprompt}. Hence, we follow prior work \cite{zhong2021factual, li-liang-2021-prefix, lester-etal-2021-power, qin-eisner-2021-learning} and model the prompts to be continuous virtual tokens instead of discrete tokens.


\subsection{Learning and Inference} 

We use teacher-forcing for model training, namely, at each step, the ground truth contexts at that step (query and previous knowledge pieces) are presented to the model. We maximize $\mathcal{L}(T)$ using standard sequence-to-sequence (seq2seq) objectives. During inference, we proceed autoregressively by feeding the recalled knowledge at step $t-1$ as the additional context at step $t$, and execute for some predefined number of steps. We also explore jointly training the prompter with a ``stopper'' which learns to stop the knowledge recall process when it decides that the recalled evidence is adequate enough; details are included in Appendix \ref{app:stopper}.

\section{Experimental Setup}
Our research question is \textit{how to shepherd a PLM to recall a series of knowledge and derive a ``chain of thought" for multi-step reasoning}. To this end, we conduct experiments on several datasets that require complex multi-step reasoning and compare different methods to guide the PLM via prompt/prefix tuning, fine-tuning, and our prompter design. We use both intrinsic and extrinsic metrics to evaluate the quality of recalled knowledge, considering both end answer accuracy and coverage of intermediate evidence.
\subsection{Datasets \& Preprocessing}
We conduct experiments on three datasets involving multi-step reasoning which include annotations for knowledge statements relevant to the queries: 2WikiMultiHopQA (abbreviated as 2Wiki) \cite{ho2020constructing}, R4C \cite{inoue-etal-2020-r4c}, and a scientific commonsense reasoning dataset (abbreviated as LoT\footnote{The abbreviation here comes from the phrase ``Leap-of-Thought'' in the paper title of \cite{talmor2020leap}.}) constructed by \cite{talmor2020leap}.


\noindent\textbf{2WikiMultiHopQA}~\cite{ho2020constructing}. 2Wiki is a recent large scale multi-hop QA dataset, which contains in total over 192k (167k train, 12.5k development, and 12.5k test) samples constructed jointly from Wikipedia and Wikidata. Since the test set is private, we randomly split the original development set into our development \& test set (6k samples each). The dataset format largely follows HotpotQA \cite{yang2018hotpotqa}, but includes more diverse reasoning types of questions and detailed annotations of evidence paths for each question. Here, an evidence path is an ordered list of (subject entity, relation, object entity) knowledge base triplets. We use the question as the query $q$, and use a simple template to convert each triplet in the evidence path into a natural language statement, forming $C_q$. Due to the large training set size and limited computing budget, we randomly sample $10\%$ of the training data to form our final training set, which has the side benefit of largely reducing the test/train overlap (more details in \S\ref{sec: faithfullness}).

\noindent\textbf{R4C}~\cite{inoue-etal-2020-r4c}. R4C is another recent multi-hop QA dataset containing annotated evidence paths. The dataset contains 4.6k examples (2.4k train, 2.2k development) constructed on top of HotpotQA, where the authors used crowd-sourcing efforts to collect the evidence paths in the form of simple subject-verb-object natural language sentences. Again, we randomly split the development set (there's no test set given) into our development and test set (1.1k samples each). We use the question as our query $q$ and use the annotated evidence sentences as $C_q$. 

\noindent\textbf{LoT} \cite{talmor2020leap}. The dataset involves reasoning over a set of taxonomic relations, constructed from ConceptNet and WordNet. Each example consists of a hypothesis (e.g., ``A whale has a belly button'') which we treat as query $q$, and a set of simple facts including hypernym rules (e.g., ``A whale is a mammal'', ``A whale is a vertebrate'') and properties (e.g., ``A mammal has a belly button'', ``A vertebrate has a tail''). By reasoning over the facts and selecting the correct chain of hypernym rule \& property (``A whale is a mammal'', ``A mammal has a belly button''), one could verify or deny the given hypothesis. One subtle issue of directly using the gold hypernym rule and property as $C_q$ is, during the first step, it would be difficult to directly identify the correct object entity without looking ahead on the properties in the second step. Therefore, for the first step, we concatenate all the hypernymic objects appearing in the dataset \textit{w.r.t.} to the same subject to form $c_1$. We drop samples from the original training set where the relevant facts are not (or only partially) provided, and obtain 9.4k/1.2k/1.2k samples for training/development/testing.

For 2Wiki and R4C, the number of steps during inference is set to be $4$ since over 99\% of the samples have less or equal to $4$ inference steps. For LoT, we set the number of inference steps to be $2$. Overall, we regard 2Wiki as our ``major'' evaluation dataset due to its largest scale (despite our down-sampling) and diverse types of queries, and use it to conduct a faithfulness study of prompting in \S\ref{sec: faithfullness}. Some examples of the processed data samples are shown in Appendix~\ref{app:egcontext}.

\subsection{Compared Methods}

We compare our proposed iterative Context-Aware Prompter (\textbf{iCAP}) along with Prompt Tuning (\textbf{Prompt-T}), Prefix Tuning (\textbf{Prefix-T}) and PLM fine-tuning (\textbf{PLM-FT}) under both non-iterative and iterative setting. The iterative setting is described in \S\ref{sec:iterprompt} and for the non-iterative setting, we simply concatenate all the knowledge statements in $C_q$ to form one single piece of knowledge for each query. In extrinsic evaluation, we also compare with fine-tuning the PLM on (query, answer) pairs without knowledge recall (\textbf{PLM-QA}), which measures how much the PLM can solve these multi-step inference problems directly, a skill which PLMs are poor at as shown by previous work. We additionally report final inference results when feeding ground truth contexts to the reader (\textbf{Oracle-RD}) as an upper bound for extrinsic evaluation. Relation-specific prompting methods such as \cite{shin-etal-2020-autoprompt, zhong2021factual, petroni2019language} are not included since they're not directly applicable to our problem setup as discussed earlier. 

Our focus in this work is on \textit{knowledge elicitation from PLMs}, and hence we do not aim to compare with previous dataset-specific methods which typically have different problem formulations \& focus than ours and utilize other attributes in the datasets which we do not use (e.g., gold \& distractor evidence paragraphs).

\begin{table*}[t]
\centering
\resizebox{0.85\textwidth}{!}{
\begin{tabular}{lcccccccc}
\toprule
     &\multicolumn{4}{c}{\textbf{2Wiki}}&\multicolumn{2}{c}{\textbf{LoT}}&\multicolumn{2}{c}{\textbf{R4C}}  \\
     \cmidrule(l{0.5em}r{0.5em}){2-5}\cmidrule(l{0.5em}r{0.5em}){6-7}\cmidrule(l{0.5em}r{0.5em}){8-9}
     & \textbf{Evi.}$\bm{R^*}$ & \textbf{Evi.}$\bm{R}$ & \textbf{Ans.}$\bm{\hat{R}}$ & \textbf{Ans.}$\bm{R}$ & \textbf{Evi.}$\bm{R^*}$ & \textbf{Evi.}$\bm{R}$ & \textbf{Ans.}$\bm{\hat{R}}$ & \textbf{Ans.}$\bm{R}$\\
     \midrule
     PLM-FT&10.3 & 33.8 & 12.3 & 45.3&41.8 & 70.8&38.1 & 43.9\\
     PLM-FT (Iter)&26.3 & 48.9 & 35.4 & 60.6 & 41.3 & 70.1 & 43.1 & 48.5 \\
     \midrule
     Prompt-T&5.5 & 22.3 & 6.6 & 41.3&35.3 & 62.8&28.2 & 33.4\\
     Prompt-T (Iter)&10.8 & 27.5 & 16.7 & 46.2&33.3 & 63.4&30.6&36.0\\
     Prefix-T&6.7 & 25.9 & 7.6 & 44.2&31.8 & 64.0&27.2 & 33.9\\
     Prefix-T (Iter)&14.8 & 33.9 & 22.5 & 53.2&31.6 & 64.9&33.7 &39.8\\
     \midrule
     iCAP&22.0 & 42.1 & 28.6 & 54.6&34.1 & 65.0&36.8 & 41.5\\
     \bottomrule
     
\end{tabular}
}
\caption{Results for Intrinsic Evaluation, where ``(Iter)'' indicates the iterative setting. All metrics are defined in \S\ref{evalmetric} and overall measure the gold (answer) entity/object coverage of the recalled knowledge from different perspectives.}
\label{table:recall}
\end{table*}
\begin{table}[t]
\centering
\resizebox{\linewidth}{!}{
\begin{tabular}{lccccc}
\toprule
     &\multicolumn{2}{c}{\textbf{2Wiki}}&\textbf{LoT}&\multicolumn{2}{c}{\textbf{R4C}}  \\
     \cmidrule(l{0.5em}r{0.5em}){2-3}\cmidrule(l{0.5em}r{0.5em}){4-4}\cmidrule(l{0.5em}r{0.5em}){5-6}
     &\textbf{EM}&\textbf{F1}&\textbf{EM}&\textbf{EM}&\textbf{F1}\\
     \midrule
     Oracle-RD&97.8 & 98.9&100.0&75.7 & 86.8\\
     PLM-QA&24.1 & 29.3&68.3&22.6 & 28.8\\
     \midrule
     PLM-FT&33.6 & 37.8&76.0&25.3 & 36.8\\
     PLM-FT (Iter)&45.5 & 50.9&77.8&32.2 & 42.5\\
     \midrule
     Prompt-T&26.9 & 31.0&65.9&16.6 & 25.9\\
     Prompt-T (Iter) &25.0 & 30.2&68.8&22.4 & 30.4\\
     Prefix-T&31.6 & 35.6&69.0&19.2 & 29.2\\
     Prefix-T (Iter) &31.1 & 36.4&72.6&24.0 & 34.2\\
     \midrule
     iCAP&42.8 & 47.9&73.8&25.7 & 35.2\\
     \bottomrule
\end{tabular}
}
\caption{Results for Extrinsic Evaluation, where the recalled knowledge of each method is used for final inference, except for Oracle-RD and PLM-QA.}
\vspace{-1em}
\label{table:qa}
\end{table}

\subsection{Evaluation Metric}
\label{evalmetric}
We use both intrinsic and extrinsic metrics to evaluate the PLM recalled knowledge.

\noindent \textbf{Intrinsic Evaluation.} Here, we directly measure the quality of recalled knowledge. While there are standard metrics for evaluating text generation such as BLEU and ROUGE, these metrics generally fail to capture the entity-centric nature of the recalled knowledge we wish to examine (more details are included in Appendix \ref{app:stdmetrics}). Therefore, we propose a set of measures that are better suited for the tasks in our experiments. 
For 2Wiki and R4C, we evaluate the ratio where the recalled knowledge contains the answer entity (\textbf{Ans.}$\bm{R}$); we also compute the ratio among only those samples where the answer entity does not appear in the query (\textbf{Ans.}$\bm{\hat{R}}$).
For 2Wiki and LoT, we also evaluate the evidence coverage of recalled contexts by computing the average ratio of gold evidence appearing in the recalled context (\textbf{Evi.}$\bm{R}$) and the ratio of samples where \textit{all} gold evidence are recalled (\textbf{Evi.}$\bm{R^*}$) as a more strict measure.
For 2Wiki, we use the entities from the annotated KB triples as evidence.
For LoT, we consider the hypernym rule/property as evidence, where in the 1st step, we deem the hypernym rule as correct if the gold object is recalled, and use exact match for the recalled property in the 2nd step.

\noindent \textbf{Extrinsic Evaluation.} We also conduct extrinsic evaluation by measuring how much the recalled knowledge help find the response to the query. Similar to reading comprehension, we concatenate all recalled knowledge as the contexts, and use a reader which tries to infer the answer given the query and contexts. For 2Wiki and R4C, we first pre-train the reader using the ground truth contexts, and then fine-tune it on the recalled contexts\footnote{We found in our preliminary experiments that this approach gives the best results across different methods.}; for LoT, we use a rule-based reader directly\footnote{LoT is constructed using templates, and therefore a rule-based reader can perfectly solve the inference task (100\% accuracy when ground truth contexts are given, see Table \ref{table:qa}).}. We report Exact Match (\textbf{EM}) and Answer \textbf{F1} scores for 2Wiki \& R4C, and \textbf{EM} score for LoT where the answer is restricted to yes/no.

\subsection{Implementation Details} 
\noindent \textbf{Architectures \& hyperparameters.} We use BART-large \cite{lewis-etal-2020-bart} for our PLM and RoBERTa-base \cite{liu2019roberta} for our prompter, which is several times smaller than the PLM\footnote{While our prompter is also initialized using a Pre-trained Language Model, we'll use the term ``PLM'' to refer only to the larger \& more knowledgeable one.}. We also include some results \& discussion for different prompter scales in Appendix \ref{app:pt_scale}. We use another BART-large for the reader in extrinsic evaluation\footnote{For the reader, we intentionally choose the same architecture with the PLM for a fair comparison with \textbf{PLM-QA}.}.


Our implementation is based on Hugging Face Transformers \cite{wolf-etal-2020-transformers}. We use AdamW optimizer \cite{loshchilov2018decoupled} and a linear learning rate scheduler with a warmup ratio of 0.06 for optimization. For hyperparameters, we use a batch size of 32, 128, 32 for 2Wiki, LoT and R4C respectively, and tune the learning rate from \{4e-5, 8e-5, 4e-4, 8e-4, 4e-3, 8e-3, 4e-2\} \& length of encoder/decoder prompts\footnote{We set the length of encoder \& decoder prompts to be the same, as we do not observe improvements otherwise in preliminary experiments.} from \{15, 30, 45, 60, 80, 100\}; more details are included in Appendix~\ref{app:hyper}. We run most experiments with three random seeds and report the mean scores.

\noindent \textbf{Knowledge Enhancement for PLM.} Since our focus is on how to make PLMs better at recalling relevant knowledge for multi-step inference, we need to make sure the PLM actually memorizes all the relevant knowledge in the first place, so that the results can be attributed solely to the effectiveness of knowledge recall. Hence, we conduct knowledge enhancement for the PLM, where we additionally pre-train the PLM to recover separately masked elements in the triplets which form the knowledge statements, a strategy similar to salient span masking \cite{roberts2020much, guu2020realm}. More details could be found in Appendix~\ref{app:ke}. Note the same PLM after knowledge enhancement is used across different compared methods.

\section{Results}

\subsection{Effectiveness of iCAP}
\label{sec:effectiveness}

The results for intrinsic \& extrinsic evaluation are summarized in Table \ref{table:recall} and \ref{table:qa} respectively, which are highly consistent. We elaborate on the results in what follows.

\noindent \textbf{Effectiveness of Iterative Scheme \& Context-Aware Prompter.} Across different datasets, it can be seen that most compared methods benefit from the iterative setting (Iter) over the non-iterative setting. Moreover, our proposed iterative Context-Aware Prompter (\textbf{iCAP}) further outperforms Prompt/Prefix Tuning by notable gains across different datasets and metrics, approaching the performance of PLM fine-tuning (\textbf{PLM-FT}); in particular, on the 2Wiki dataset which has the largest scale and diversity of reasoning types, \textbf{iCAP} achieves more than 15\% and 10\% absolute gains in F1 over Prompt-Tuning \& Prefix-Tuning respectively. Overall, the results clearly show the effectiveness of the iterative scheme and our proposed context-aware prompter design. However, we note that even the best results (prompting based or fine-tuning based) still far lag behind \textbf{Oracle-RD} which uses ground truth contexts as input, which suggests a large room for improvements with better methods for knowledge elicitation from PLMs. Some failure cases of iCAP are included in Appendix~\ref{app:egcontext}.

\noindent \textbf{Helpfulness of Knowledge Recall for Multi-step Inference}. The result obtained by fine-tuning the PLM on (query, answer) directly without knowledge recall (\textbf{PLM-QA}) is outperformed by almost all other compared methods, verifying the previous findings that PLMs face difficulties in using their stored knowledge to perform multi-step inference tasks. The large gain obtained from methods based on knowledge recall shows the helpfulness of deriving a ``chain of thought'' (especially iteratively) from PLMs for multi-step inference.



\begin{table*}[h]
\centering
\scalebox{0.95}{
\begin{tabular}{lcccccccc}
\toprule
&\multicolumn{4}{c}{\textbf{Random Model}}&\multicolumn{4}{c}{\textbf{Random Embedding}} \\
\cmidrule(l{0.5em}r{0.5em}){2-5}\cmidrule(l{0.5em}r{0.5em}){6-9}
&\textbf{Evi.}$\bm{R^*}$ & \textbf{Evi.}$\bm{R}$ & \textbf{Ans.}$\bm{\hat{R}}$ & \textbf{Ans.}$\bm{R}$&\textbf{Evi.}$\bm{R^*}$ & \textbf{Evi.}$\bm{R}$ & \textbf{Ans.}$\bm{\hat{R}}$ & \textbf{Ans.}$\bm{R}$ \\
\midrule
PLM-FT&1.77 & 5.20 & 3.76 & 37.48&4.10 & 11.47 & 6.52 & 37.18	\\
Prompt-T&0.0 & 0.0 & 0.0 & 0.0&0.006 & 0.013 & 0.003 & 0.002	 \\ 
Prefix-T&0.001 & 0.0 & 0.0 & 0.0&0.009 & 0.014 & 0.004 & 0.002		 \\ 
iCAP&0.001 & 0.001 & 0.0 & 0.0&1.49 & 2.83 & 0.98 & 0.59		 \\
\bottomrule
\end{tabular}}
\caption{Intrinsic Evaluation Results on Random Control Experiments. Here we only focus on the iterative setting using the 2Wiki dataset.}
\label{table:randctrl}
\end{table*}

\subsection{Faithfulness of Prompting}
\label{sec: faithfullness}
\cite{zhong2021factual} raised and studied some important questions in optimization-based prompting methods: \textit{Are the prompts ``really'' doing prompting? Is it possible that they capture dataset regularities too?} The issue is related to the notion of test-train overlap \cite{lewis-etal-2021-question}, where the dataset may contain some underlying spurious patterns that the model exploits, and thus standard evaluations could not truthfully measure their generalization behaviors. Here, we take this concern seriously and conduct a series of analysis to faithfully interpret the results we obtained. We focus on 2Wiki under iterative setting for our analysis.

\noindent \textbf{Test-Train Overlap.} For each development \& test sample, we compute the ratio of knowledge statements in $C_q$ that also appear in the training set, meaning that during certain steps for some training samples, the model has ``seen'' the exact same piece of knowledge. Note that this is a rather strict measure: even if all the knowledge pieces in $C_q$ are seen during training, they may come from completely different samples \& steps and hence organized in different ways. We summarize the overlapping ratios of development \& test set samples in Table \ref{table:overlap} in Appendix~\ref{app:overlap}. It can be seen that the down-sampling has the side benefit of greatly reducing the test-train overlap; in particular, the percentage of examples where all knowledge statements are seen during training is reduced from almost 30\% to less than 2\%, and more importantly, over 70\% of the samples have no overlap. This suggests a rather low risk for the existence of strong spurious regularities in our setup.

\noindent \textbf{Random Control Experiments.} Examining the data-level statistics is helpful, but still not sufficient in terms of revealing the spurious regularities that different methods may capture. Hence, we follow \cite{zhong2021factual} to conduct two random control experiments. In the \textit{Random Model} experiment, we re-initialize all parameters of the PLM to clean out its internal knowledge, and proceed with the same training procedure as earlier. In this way, any positive signal obtained could only be attributed to dataset regularities captured by the method. In the \textit{Random Embedding} experiment, we re-initialize only the input embeddings of the PLM, a setting analogous to the \textit{control task} introduced in \cite{hewitt2019designing} (more discussions can be found in \cite{zhong2021factual}). Here we only proceed with the iterative setting and conduct intrinsic evaluation, where the results are summarized in Table \ref{table:randctrl}. It can be seen that 1) PLM fine-tuning captures significantly more regularities in the dataset than prompting-based methods; 2) While our proposed method captures a bit more regularities than Prompt/Prefix Tuning, they still remain at a very small level. Overall, our random control experiments show that the exploitation of spurious dataset patterns by the evaluated prompting methods is rather mild, and that by PLM fine-tuning could potentially be larger.

\noindent\textbf{Prompter Attention Visualization.} To see whether our proposed iCAP behaves in the way we expect, one direct approach is to examine the inner workings of the prompter. Towards this end, we visualize the attentions during the prompter forward pass at different steps. We randomly choose examples in the development/test set, and use BertViz \cite{vig-2019-multiscale} to visualize the attentions within the forward pass of the prompter after the following processing steps: 1) we aggregate the attention weights of different attention heads within the same transformer layer; 2) to better view the prompt tokens as one single unit, we average the attentions across different prompt tokens to form one ``master'' prompt token; 3) we drop all special tokens (BOS, EOS) for cleaner visualization. One example (the same example which we use in Figure \ref{fig:pipeline}) is in Figure \ref{fig:attn}, and we include more examples in Appendix~\ref{app:example}. As briefly illustrated earlier in \S\ref{sec:intro}, during the 1st step towards solving this query, the prompter should focus on the part concerning ``father'' of ``Gwilym Lloyd George''; during the 2nd step, the prompter should integrate the answer ``David Lloyd George'' from the 1st step evidence and the ``place of birth'' part in the query to synthesize the prompt. We can see that the attention distributions at different steps accord well with our expectations. However, we note that attention visualization is only a qualitative approach; more systematic ways for examining the inner working behaviors of transformers remain an open challenge.

\begin{figure}[t]
  \centering
    \includegraphics[width=0.48\textwidth]{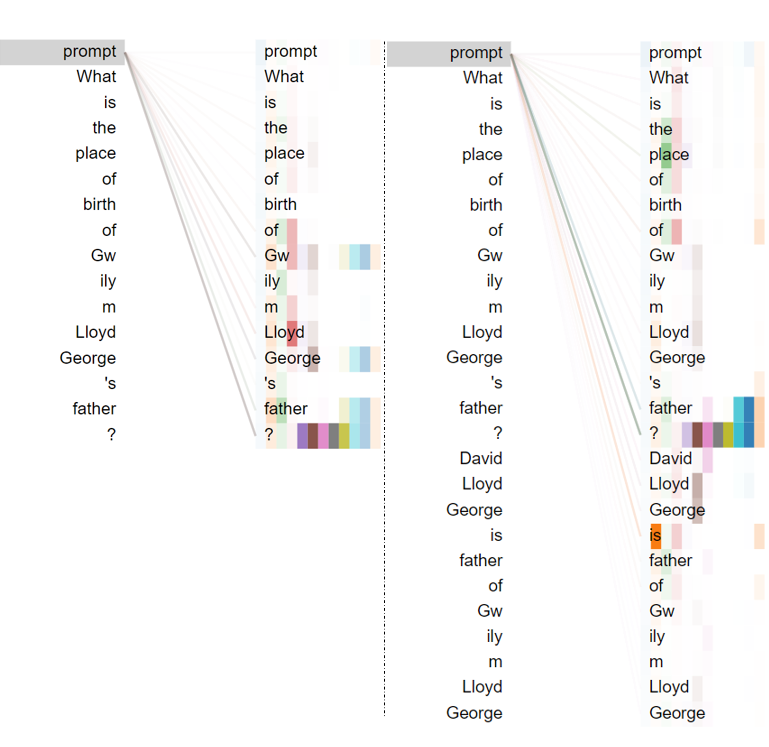}
  \caption{Prompter Attention Visualization. Attentions during the forward pass for the 1st \& 2nd step are shown on the left \& right respectively. Different colors correspond to different transformer layers. More examples of different reasoning types are included in Appendix~\ref{app:example}.}
\label{fig:attn}
\end{figure}

\section{Related Work}
\noindent\textbf{Memorization and Reasoning in PLMs.}
With the recent success of large-scale pre-trained language models (PLMs), there has been growing interest in investigating what is captured by these PLMs during pre-training~\cite{talmor2020olmpics, rogers2020primer, kassner2020pretrained}. Studies have shown that in addition to learning linguistic knowledge about language use, PLMs are capable of memorizing a great amount of world knowledge~\cite{rogers2020primer}, yielding competitive performance on knowledge probing \cite{petroni2019language, shin-etal-2020-autoprompt, zhong2021factual} and other knowledge-intensive tasks such as question answering \cite{roberts2020much} and fact checking \cite{lee2020language}, without resorting to any external knowledge source. On the other hand, other work such as \cite{talmor2020olmpics, kassner2020pretrained, rae2021scaling} reveals that PLMs face difficulties in recalling their stored knowledge for \emph{multi-step} inferences (such as answering complex, multi-hop questions), which is also verified in our experiments.

\noindent\textbf{Prompt Learning.} One type of method for eliciting knowledge from PLMs is prompting \cite{liu2021pre}, which is gaining increasing research interests \& potential recently. Prompting methods seek to re-frame queries into prompts which accord with the PLM's input format, and extract useful information from the predicted results. The benefit of not needing to tune PLMs makes prompting especially appealing as PLMs scale up in size. In this work, we are interested in developing prompting methods which could enable PLMs to recall a series of relevant knowledge for multi-step inference. Previous work along this direction mainly use manually designed prompts/templates suited for specific datasets \cite{paranjape-etal-2021-prompting, mishra2021reframing, shwartz-etal-2020-unsupervised}; instead, we seek to develop a general method which can learn to construct appropriate prompts automatically. Concurrent to our work, Chain-of-Thought (CoT) prompting \cite{wei2022chain} shares similar high-level ideas as ours, where the authors propose to provide intermediate reasoning steps in the prompts to encourage the PLM to perform step-by-step inference. While CoT shows great successes, we note it is one of the \textit{emergent abilities} of large language models \cite{wei2022emergent} and only works well with extremely large PLMs (>100B typically) such as GPT-3 \cite{brown2020language} and PaLM \cite{chowdhery2022palm}. In our work, we use PLMs
that are several orders of magnitude smaller than those used in CoT and demand much less computing resources. We hope our efforts could contribute towards developing LM-based systems with better multi-step reasoning abilities but also moderate scale.

For existing work on learning-based prompting, \cite{shin-etal-2020-autoprompt} proposes to use gradient-guided search to find appropriate discrete prompt tokens in the PLM's vocabulary to form prompt templates. While the resulting prompts are readable, most of them have very low fluency and interpretability. \cite{zhong2021factual, qin-eisner-2021-learning} propose to optimize the prompts in continuous space instead, which shows large benefits in both effectiveness and optimization efficiency. \cite{zhong2021factual} also raises and studies the question of whether learning-based prompting could exploit spurious dataset regularities which would weaken the validity of standard evaluation results, a concern we seriously address in our work. \cite{lester-etal-2021-power, li-liang-2021-prefix} follow the continuous prompting paradigm, and tune task-level prompts for lightweight adaptation of PLMs. Overall, existing prompt learning methods are either restricted to cases where there exists a single \& identifiable relation/predicate within the query \cite{zhong2021factual, qin-eisner-2021-learning, shin-etal-2020-autoprompt}, or being static and not sensitive to sample-wise inputs \cite{lester-etal-2021-power, li-liang-2021-prefix}.

\noindent \textbf{Iterative Knowledge Retrieval.} We are also inspired by methods that iteratively retrieve knowledge from explicit knowledge sources for multi-step reasoning, such as \cite{xiong2021answering, qi-etal-2019-answering, khattab2021baleen, mo2022knowledge}. Our problem setting could be viewed as iterative retrieval over implicit knowledge in PLMs, instead of from explicit knowledge sources.

\section{Conclusion \& Future Work}
We explore an iterative prompting framework towards driving a ``chain of thought'' from PLMs for multi-step reasoning tasks. We show the superiority of this iterative scheme, and also the effectiveness of our proposed context-aware prompter design, which addresses key limitations of previous prompting methods when applied in this new scheme. In addition, we conduct both quantitative \& qualitative analysis on the faithfulness of the learned prompting behaviors. In the future, we aim to further extend and apply our ideas to language model pretraining, with the hope that PLMs can be inherently equipped with stronger multi-step reasoning capabilities. The iterative framework we explore here also opens the possibility of human intervention and interaction during inference; namely, a human can track along the PLM's chain of thought and make edits and corrections at different steps, similarly as in \cite{mo2022towards}, which improves the transparency and trustworthiness of inference and also helps reduce error propagation along the reasoning process. We leave these investigations as future work.


\section*{Limitations}
\noindent \textbf{Experiments with larger-scale models.} We explored a novel framework to prompt (or elicit knowledge from) PLMs for multi-step inference.
Although our iterative prompting approach outperforms the baselines by a large margin, there is still much room to improve. One promising direction is to experiment with PLMs larger than what is used in our experiments (i.e., BART-large), which have better capacities for internalizing knowledge. However, when the models get larger, the associated computational cost will increase accordingly, which was also the main obstacle for us to pursue this front. We intend to conduct such experiments in the future when we have access to better computing resources.



\noindent \textbf{Datasets with noisy knowledge statements.} We used three recently released datasets (2Wiki, R4C, LoT) that require multi-step inference for our experiments. Compared with alternative datasets such as HotpotQA and StrategyQA \cite{geva2021did}, they include knowledge statements that have cleaner formats and are much more suitable for multi-step inference (in fact, this is one of the main motivations behind the construction of 2Wiki \& R4C). For HotpotQA \& StrategyQA, the knowledge statements are given as raw sentences from the evidence paragraphs and include information irrelevant to the original question. We exercised our best effort to process them (e.g., resolving coreferences, simplifying \& decomposing nested sentences, etc.) into our desired formats, but the resulting knowledge statements are still very noisy. All methods including ours cannot be trained well under such knowledge statements. How to use such naturally occurring but noisy knowledge statements as supervision to guide PLMs to develop a chain of thought is an interesting topic to study in the future.


\noindent \textbf{Exploring alternative architectural designs.} Another limitation is that we only implemented an intuitive and simple instantiation (Figure 2) of our proposed context-aware prompter to illustrate its promises. It is an interesting future direction to further explore various design choices for iterative prompting, e.g., alternative design for the Prompter-PLM interface, dynamic prompt length across different inference steps, etc.

\section*{Acknowledgements}
The authors would like to thank colleagues from the OSU NLP group for their thoughtful comments. This research was supported in part by Google Faculty Award, Google Research Scholar Award, NSF IIS 1815674, NSF CAREER 1942980, and Ohio Supercomputer Center~\cite{OhioSupercomputerCenter1987}.

\balance
\bibliography{anthology,custom}
\bibliographystyle{acl_natbib}

\appendix
\newpage
\nobalance
\section{Appendix}
\subsection{Additional Details on Experiments}
\label{app:hyper}
\begin{table}[t]
\centering
\resizebox{\linewidth}{!}{
    \begin{tabular}{lcccccc}
    \toprule
    &\multicolumn{2}{c}{\textbf{2Wiki}}&\multicolumn{2}{c}{\textbf{LoT}}&\multicolumn{2}{c}{\textbf{R4C}} \\
    \cmidrule(lr){2-3}\cmidrule(lr){4-5}\cmidrule(lr){6-7}
    &\textbf{lr} & \textbf{pt\_len}&\textbf{lr} & \textbf{pt\_len}&\textbf{lr} & \textbf{pt\_len} \\
    \midrule
    Prompt-T&8e-3 & 80&4e-3 & 80&4e-3 & 60 \\ 
    Prefix-T&8e-4 & 80&4e-4 & 60&4e-4 & 80 \\ 
    PLM-FT&4e-5 & -	&4e-5 & -	&4e-5 & -	\\
    PLM-QA&4e-5 & -	&8e-5 & -	&4e-5 & -	\\
    iCAP&8e-5 & 30 &8e-5 & 60 &8e-5 & 30 \\
    \bottomrule
    \end{tabular}
    
    }
\caption{Hyperparameter settings for all compared methods. lr: learning rate, pt\_len: prompt length.}
\label{table:hyperparam}
\end{table}
\begin{table}[t]
\centering
\resizebox{0.8\linewidth}{!}{
\begin{tabular}{lcccc}
\toprule
     &\multicolumn{3}{c}{\textbf{Rouge (R)}}&\multirow{2}{*}{\textbf{BLEU}} \\
     \cmidrule(l{0.5em}r{0.5em}){2-4}
     & \textbf{R-1} & \textbf{R-2} & \textbf{R-L} &\\
     \midrule
     PLM-FT&74.3 & 62.4 & 72.7 & 52.9\\
     PLM-FT (Iter)& 83.6 & 76.3 & 82.3 & 70.8 \\
     \midrule
     Prompt-T& 68.7 & 55.5 & 66.4 & 45.4\\
     Prompt-T (Iter)&74.5 & 64.7 & 73.7 & 56.7\\
     Prefix-T&70.8 & 57.8 & 68.9 & 48.7\\
     Prefix-T (Iter)&79.0 & 70.3 & 77.6 & 64.0\\
     \midrule
     iCAP&79.2& 70.5 & 78.3 & 64.9\\
     \bottomrule
\end{tabular}
}
\caption{Intrinsic evaluation on 2Wiki using standard text generation metrics (ROUGE \& BLEU).}
\label{table:stdmetrics}
\end{table}
\begin{table}[t]
\centering
\resizebox{\linewidth}{!}{
\begin{tabular}{lcccc}
\toprule
     & \textbf{Evi.}$\bm{R^*}$ & \textbf{Evi.}$\bm{R}$ & \textbf{Ans.}$\bm{\hat{R}}$ & \textbf{Ans.}$\bm{R}$\\
     \midrule
     iCAP & 20.0 & 39.1 & 26.5 & 54.0 \\
     iCAP (with stopper) &18.4& 37.5 & 22.9 & 51.8\\
     \bottomrule
\end{tabular}
}
\caption{Intrinsic evaluation results from jointly training the Prompter and the Stopper which learns to stop the knowledge recall process when it decides that the recalled knowledge is adequate enough for answering the query.}
\label{table:stopper}
\end{table}
\begin{table*}[t]
\centering
\scalebox{0.95}{
\begin{tabular}{lccccccc}
\toprule
&0\% & 1\%-20\% & 21\%-40\% & 41\%-60\% & 61\%-80\% & 81-99\% & 100\%\\
\midrule
\textbf{2wiki (full)}&36.0\% & 0.0\% & 0.5\% & 28.4\% & 5.2\% & 0.0\% & 29.8\%\\
\textbf{2wiki (down-sampled)}&71.4\% & 0.1\% & 8.1\% & 16.2\% & 2.6\% & 0.0\% & 1.6\%\\
\bottomrule
\end{tabular}}
\caption{Test/Train simple knowledge overlap on 2Wiki. The horizontal bar represents the percentage range of simple knowledge statements appearing in the training set, and the content values are the percentages of development \& test set examples that fall into the corresponding range.} 
\label{table:overlap}
\end{table*}
\begin{table*}[t]
\centering
\scalebox{1.0}{
\begin{tabular}{cccccccc}
\toprule
\multicolumn{4}{c}{\textbf{BERT-tiny}}&\multicolumn{4}{c}{\textbf{BERT-small}} \\
\cmidrule(l{0.5em}r{0.5em}){1-4}\cmidrule(l{0.5em}r{0.5em}){5-8}
\textbf{Evi.}$\bm{R^*}$ & \textbf{Evi.}$\bm{R}$ & \textbf{Ans.}$\bm{\hat{R}}$ & \textbf{Ans.}$\bm{R}$&\textbf{Evi.}$\bm{R^*}$ & \textbf{Evi.}$\bm{R}$ & \textbf{Ans.}$\bm{\hat{R}}$ & \textbf{Ans.}$\bm{R}$ \\
\midrule
6.0 & 17.7 & 9.0 & 35.3 & 21.4 & 41.2 & 29.1 & 54.2	\\
\bottomrule
\end{tabular}}
\caption{2Wiki intrinsic evaluation results with two smaller-scale prompter instantiations.}
\label{table:prompter_scale}
\end{table*}
\textbf{Hyperparameters.} We set the batch size to be 32, 128, 32 and train for 70, 50, 40 epochs for 2Wiki, LoT \& R4C respectively. Table \ref{table:hyperparam} summarizes other hyperparameters used in our experiments.

\subsection{More details on PLM Knowledge Enhancement}
\label{app:ke}
To make sure the PLM knows all the relevant knowledge for subsequent recall, we further pre-train the PLM to recover separately masked elements in the triplets which form the knowledge statements. For 2Wiki and LoT, we also additionally include knowledge statements that are not used in the dataset to make the setting more challenging; one can think of these extra knowledge statements as ``distractors''. For 2Wiki, we filter from the processed Wikidata triples provided by \cite{agarwal-etal-2021-knowledge} by keeping those with subject entities appearing in the original knowledge statements, and in the end, we obtain 383k extra knowledge statements v.s. 240k original ones (note that while we downsample the training set during our main experiment, the knowledge enhancement step is performed on the full dataset). For LoT, we directly use the provided distractor knowledge in the original dataset. We don't add distractors for R4C because the provided knowledge statements are in natural language and it's hard to retrieve high quality knowledge statements as such. We verified that the PLM after knowledge enhancement can indeed recover the masked elements in the knowledge statements in near-perfect accuracy.

\subsection{Standard Metrics for Intrinsic Evaluation}
\label{app:stdmetrics}
The intrinsic evaluation results obtained by using standard text generation metrics (ROUGE 1/2/L \& BLEU) for 2Wiki are shown in Table \ref{table:stdmetrics}. Comparing with results using our proposed metrics (Table \ref{table:recall}), it could be seen that while overall they show the same trend, the standard evaluation results tend to group closer due to their lack of focus on the important parts (e.g., entities) of the recalled evidence.

\subsection{Prompter with Automatic Stopping}
\label{app:stopper}
Here we explore augmenting an additional Stopper module which could learn to decide to stop the knowledge recall process appropriately when the recalled evidence pieces are enough to answer the query. Since the representations from the Prompter are already rich, we design the Stopper module to be a simple feed-forward DNN on top of the [CLS] embedding of the Prompter. The DNN has two hidden layers of dimensions $500$ and $100$ respectively, and outputs the probability of stopping the knowledge recall process. The loss for the Stopper is standard binary classification loss, which is combined with the original Prompter loss with weight factor $0.1$. The Prompter and Stopper are jointly trained under this combined objective.

We experiment on 2Wiki only and run the experiment once due to efficiency considerations. We first evaluate the frequency that the Stopper decides to stop the recall at the same number of steps as in the ground truth knowledge pieces. Note that this is not a perfect measure, as the actual recalled knowledge is different from the ground truth knowledge. The frequency is 98.5\%, which indicates that the stopper can learn to stop the recall process appropriately. Then we use our intrinsic measures to see the quality of the recalled evidence after truncation by the Stopper; the results are shown in Table \ref{table:stopper}. Note that here, the ``iCAP'' setting (top row) is different from that in Table \ref{table:recall} (despite having the same name) since the prompter is trained together with the stopper for fair comparison. It can be seen from the results that there're small performance drops after truncating by the Stopper, which suggests that the Stopper can learn to stop the knowledge recall process rather appropriately but not perfectly.
\subsection{Test-Train Overlap}
\label{app:overlap}
Table~\ref{table:overlap} shows the 2Wiki Test-Train knowledge statement overlap, where \textbf{2Wiki (full)} corresponds to the statistics using the full training set, and \textbf{2Wiki (down-sampled)} corresponds to the down-sampled training set that we used in our actual experiment. The inference steps in 2Wiki are mostly 2 or 4, so overall there're higher chances for the coverage ratio to be 50\%.

\subsection{Examples of processed data samples \& failure cases of iCAP}
\label{app:egcontext}
Table~\ref{tab:example_contexts} shows examples of our processed data samples for each dataset and each sub-category, along with some failure cases of our proposed method.
\subsection{Variants of Prompter Scales}
\label{app:pt_scale}
While we used RoBERTa-base to instantiate the prompter in our main experiments, it is also interesting to see how the performance varies along different scales of the prompter. Towards this end, we conducted experiments on 2Wiki with two smaller scale prompters: BERT-small (28.8 million parameters) \& BERT-tiny (4.4 million parameters). The intrinsic evaluation results are shown in Table~\ref{table:prompter_scale}. It can be seen that the performance grows as the prompter scale grows; in addition, BERT-small can also achieve an impressive performance (under-performing RoBERTa-base used in our main experiments by just a small gap) while BERT-tiny basically fails. This suggests that the prompter still needs to be larger than a certain scale for our method to work well. 

\subsection{More Examples on Prompter Attention Visualizations}
\label{app:example}
\begin{figure}[h]
  \centering
    \includegraphics[width=0.48\textwidth]{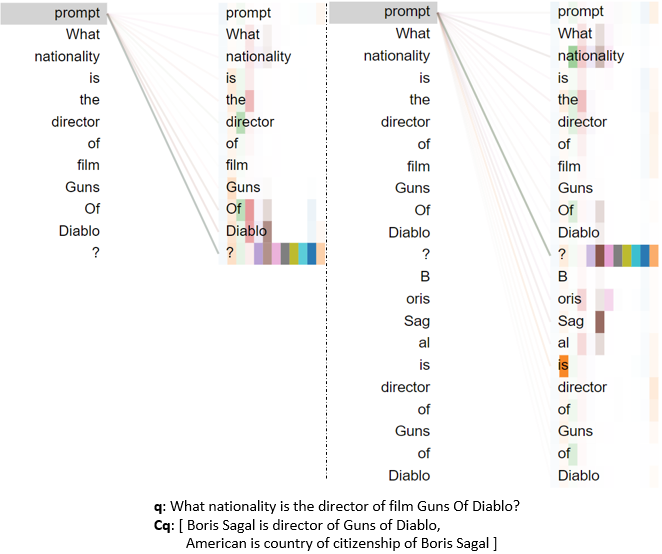}
  \caption{Prompter attention visualization. Reasoning type: Composition.}
\label{fig:prompter_1}
\end{figure}

\begin{figure}[t]
  \centering
    \includegraphics[width=0.48\textwidth]{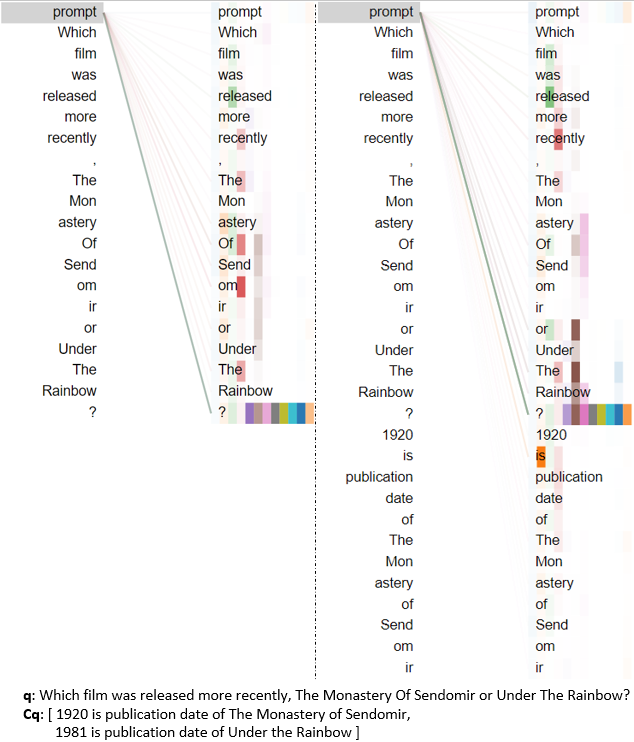}
  \caption{Prompter attention visualization. Reasoning type: Comparison.}
\label{fig:prompter_2}
\end{figure}

\begin{figure}[t]
  \centering
    \includegraphics[width=0.48\textwidth]{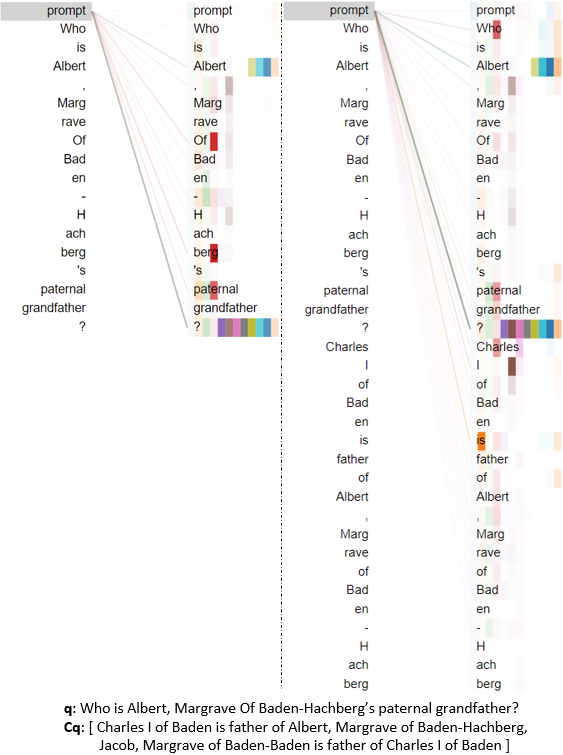}
  \caption{Prompter attention visualization. Reasoning type: Inference.}
\label{fig:prompter_3}
\end{figure}

\begin{figure*}[t]
  \centering
    \includegraphics[width=1.0\textwidth]{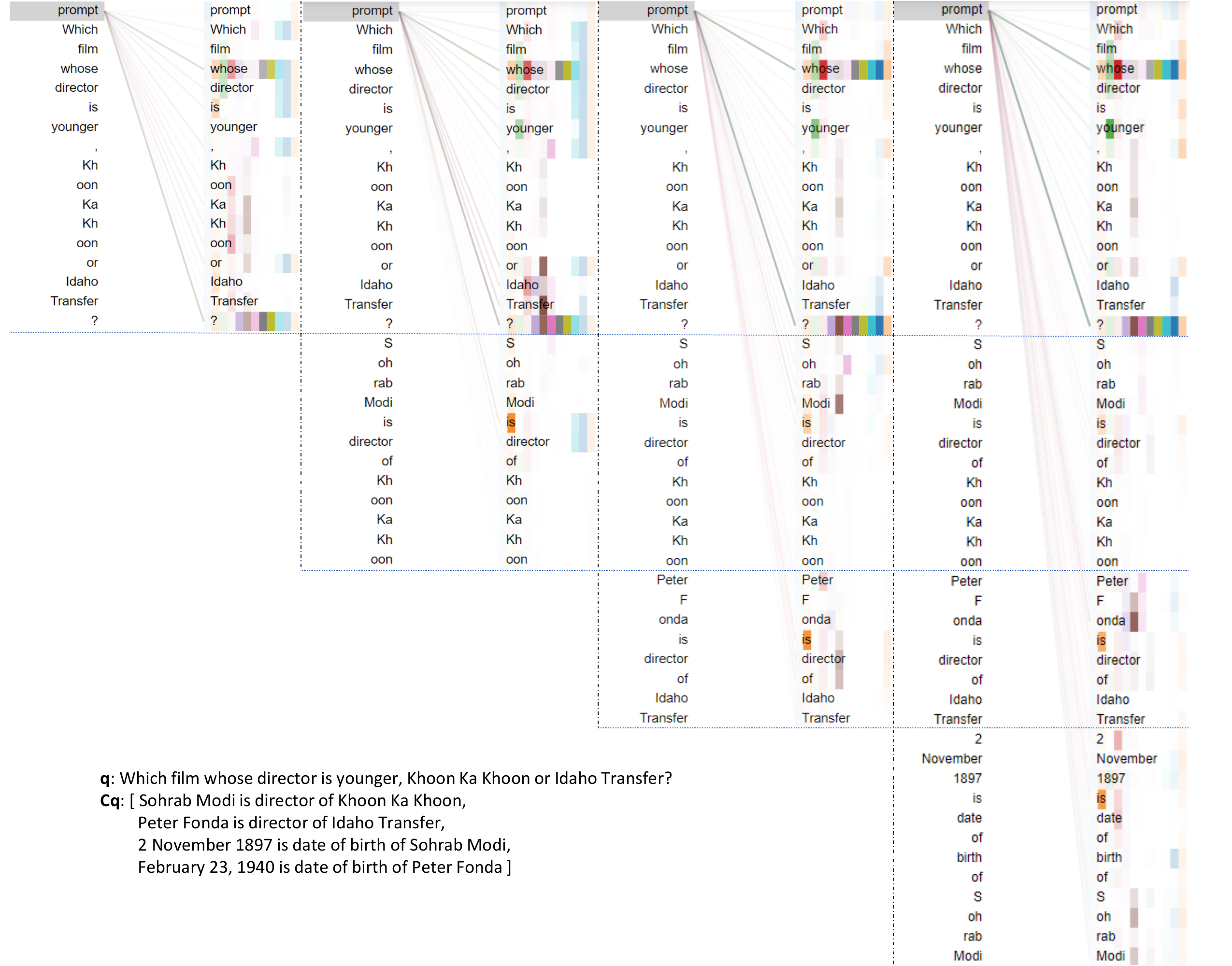}
  \caption{Prompter attention visualization. Reasoning type: Bridge-comparison.}
\label{fig:prompter_4}
\end{figure*}

Figure \ref{fig:prompter_1}, \ref{fig:prompter_2}, \ref{fig:prompter_3}, \ref{fig:prompter_4} show additional example prompter attention visualizations in the 2Wiki dataset, each corresponding to a different reasoning type as indicated in the captions.

\pagebreak

\onecolumn
\pagebreak
\twocolumn

\begin{table*}[]
    \centering
    \resizebox{\linewidth}{!}{
    \begin{tabular}{lp{12cm}}
       \toprule
        Query (2Wiki[Composition])  &  What is the place of birth of the performer of song La Terre Est Ronde?\\
        \hdashline
        Gold Knowledge &  Orelsan is performer of La terre est ronde; Alençon is place of birth of Orelsan\\ 
        \hdashline
        Recalled Knowledge & Basshunter is performer of La Terre est ronde; Havana is place of birth of Basshunter \\
       
       \toprule
        Query (2Wiki[Comparison]) & Who was born first out of Emma Kealy and Viktor Podloucký?\\
        \hdashline
        Gold Knowledge & 29 May 1977 is date of birth of Emma Kealy; December 3, 1950 is date of birth of Viktor Podloucký \\
        \hdashline
        Recalled Knowledge & 30 March 1977 is date of birth of Emma Kealy; 9 October 1964 is date of birth of Viktor Podloucký \\

       \toprule
        Query (2Wiki[Inference])  &  Who is the maternal grandfather of Vyacheslav Yaroslavich?\\
        \hdashline
        Gold Knowledge & Ingegerd Olofsdotter of Sweden is mother of Vyacheslav Yaroslavich; Olof Skötkonung is father of Ingegerd Olofsdotter of Sweden\\ 
        \hdashline
        Recalled Knowledge & Yaroslavlava of Avidia is mother of Vyacheslav Yaroslavich; Sovatoslav is father of Yaroslavlava of Avidia \\

       \toprule
        Query (2Wiki[Bridge comparison]) & Which film has the director died later, One Day In The Life Of Andrei Arsenevich or Wolves Of The Range?\\
        \hdashline

        Gold Knowledge & Chris Marker is director of One Day in the Life of Andrei Arsenevich; Sam Newfield is director of Wolves of the Range; 29 July 2012 is date of death of Chris Marker; November 10, 1964 is date of death of Sam Newfield\\
        \hdashline
        Recalled Knowledge & Chris Marker is director of One Day in the Life of Andrei Arsenevich; Wallace Fox is director of Wolves of the Range; 21 January 2013 is date of death of Chris Marker; March 30, 1999 is date of death of Andrei Arsenevich\\

        \toprule
        Query (LoT) & A evergreen is a important food source.\\
        \hdashline

        Gold Knowledge & A evergreen is a plant; A plant is not a important food source\\ 
        \hdashline
        Recalled Knowledge & A evergreen is a material, tree; A tree is a important food source\\
     
        \toprule
        Query (R4C[Comparison])  &  Which documentary was filmed first, Almost Sunrise or Hail! Hail! Rock 'n' Roll?\\
        \hdashline

        Gold Knowledge &  Almost Sunrise was filmed in 2016; Hail! Hail! Rock 'n' Roll was filmed in 1986\\
        \hdashline
        Recalled Knowledge & Almost Sunrise (album) is credited to American singer-songwriter Taylor Swift; Rock 'n' Roll is filmed in the 1990s\\
        
        \toprule
        Query (R4C[Bridge]) & Who was the chief executive officer of the second largest US car rental company by sales?\\
        \hdashline

        Gold Knowledge & The Hertz Corporation is the second-largest US car rental company; Robert L. Stone was chief executive officer of The Hertz Corporation\\
        \hdashline
        Recalled Knowledge & The Hertz Corporation is the second-largest US car rental company; Enterprise Rent-A-Car founder Jack Taylor was chief executive officer of Hertz\\

        \bottomrule
    \end{tabular}}
    \caption{Examples of our processed data samples for each dataset and sub-category (indicated in brackets), along with failure cases of our method.}
    \label{tab:example_contexts}
\end{table*}

\end{document}